\documentclass{article}
\usepackage{ijcai20}

\usepackage{times}
\usepackage{soul}
\usepackage{url}
\usepackage[hidelinks]{hyperref}
\usepackage[utf8]{inputenc}
\usepackage[small]{caption}
\usepackage{graphicx}
\usepackage{amsmath}
\usepackage{amsthm}
\usepackage{booktabs}
\usepackage{algorithm}
\usepackage{algorithmic}
\usepackage{adjustbox}
 \usepackage{multirow}
\usepackage{amssymb}
\usepackage{bbm}
\usepackage{amsmath}
\usepackage{mathtools}
\usepackage{graphicx}
\usepackage{subcaption}
\usepackage{caption}
\usepackage{microtype}
\usepackage{booktabs} 
\usepackage{mathrsfs,epsfig,epstopdf}
\usepackage[algo2e,algonl,ruled]{algorithm2e}
\usepackage{tikz}
\usepackage{pgf}
\usepackage{float}
\usepackage[bottom]{footmisc}
\usepackage{array}
\usepackage{hyperref}
\usepackage{appendix}
\usepackage{bm}

\newcommand\numberthis{\addtocounter{equation}{1}\tag{\theequation}}
\newcommand{\R}{\mathbb{R}}
\newcommand{\x} {\mathbf{x}}

\def \I {\mathbb{I}}

\newtheorem{theorem}{Theorem}

\title{Robust Deep Ordinal Regression under Label Noise}

\author{
Bhanu Garg$^1$
\and
Naresh Manwani$^2$
\affiliations
$^{1,2}$Machine Learning Lab, IIIT Hyderabad, India\\
\emails
bhanugarg05@gmail.com,
naresh.manwani@iiit.ac.in
}

\begin{document}

\maketitle
\begin{abstract}
The real-world data is often susceptible to label noise, which might constrict the effectiveness of the existing state of the art algorithms for ordinal regression. Existing works on ordinal regression do not take label noise into account. We propose a theoretically grounded approach for class conditional label noise in ordinal regression problems. We present a deep learning implementation of two commonly used loss functions for ordinal regression that is both - 1) robust to label noise, and 2) rank consistent for a good ranking rule. We verify these properties of the algorithm empirically and show robustness to label noise on real data and rank consistency. To the best of our knowledge, this is the first approach for  robust ordinal regression models. 
\end{abstract}

\section{Introduction}
\noindent Ordinal regression, or sometimes ranking learning, is a supervised learning problem where the objective is to predict categories or labels on an ordinal scale. Ordinal regression frequently arises in social sciences and information retrieval, where human preferences play a significant role. The label space does not have a distance metric defined over it, which distinguishes it from regression problems, and the relative ordering among the labels distinguishes it from multiclass classification.



Common applications of ordinal regression include age detection from face images, predicting credit ratings \cite{credit_ratings}, progress of diseases such as Alzheimer's \cite{alzeimhers}, periodontal diseases \cite{diseases}, decoding information on neural activity from fMRI scans \cite{decoding_brain} to name a few. Such varied and high impact applications make ordinal regression  an important learning model.  

An ordinal regression is commonly described by a real-valued function and a set of ordered thresholds. Many state-of-the-art methods in supervised learning use risk-minimization techniques to learn the model, which requires a suitable loss function. Commonly used  zero-one loss for classification problems would ignore the ordinal nature of the labels.  Instead, mean absolute error (MAE), defined as the absolute difference between the ranks of the predicted and the true label, is used to evaluate the performance of ordinal regression approaches. However, MAE is not  continuous, which makes risk minimization computationally hard. As a consequence, convex surrogates of MAE are used for risk minimization. One such loss function is the implicit constrained loss ($l_{IMC}$) proposed in \cite{Chu2005NewAT}. It is used to learn maximum margin ordinal regression function \cite{Chu2005NewAT,Antoniuk2016}. Perceptron based online ranking algorithms are proposed in \cite{Crammer:2001:PR:2980539.2980623,DBLP:pril/corr/abs-1802-03873}. The $l_{IMC}$ and the above online algorithms preserve the ordering of thresholds on risk minimization. Ordering of thresholds can also be forced by posing the constraints explicitly \cite{Chu2005NewAT}. \cite{Li:2006:ORE:2976456.2976565} propose an approach that converts ordinal regression learning into extended binary classification. Neural networks have also been used to learn ordinal regression \cite{DBLP:journals/corr/abs-1901-07884,DBLP:nn_based_approach}. In \cite{DBLP:journals/corr/abs-1901-07884}, authors use cross entropy-based loss ($l_{CE}$) for ordinal regression and show that $l_{CE}$ intrinsically maintains the ordering among the thresholds. A deep neural network model for ordinal regression is proposed in \cite{Liu2018ACD}. All the works above assume that the data used for the training does not suffer from label noise.

Because of practical constraints with the way data is collected, the labels in the data might be noisy. Subjective errors, measurement errors, manual errors etc. are some of the reasons we get noisy labels. Because of this label noise in the data, we may not learn the correct underlying ordinal regression function. Thus, we need to develop robust methods that can learn the actual underlying classifier even when we have label noise in the training data. 

Label noise problems in the context of binary and multiclass classification problems is an active area of research. A thorough literature survey of label noise robust methods for classification is provided in \cite{noise_survey}. In \cite{DBLP:noise_tolerance_nar,ghosh2017robust}, authors provide sufficient symmetry conditions on loss functions that would ensure robustness to label noise for classification. It is shown in \cite{DBLP:noise_tolerance_nar,ghosh2017robust} that convex loss functions are not robust under label noise for binary classification. Similar results are shown by \cite{ghosh2017robust} for multiclass classification. On the other hand, the approach in \cite{Natarajan:2013:LNL:2999611.2999745} assumes the knowledge of noise rates and finds an unbiased estimator of the true risk under noisy labels. The authors also show that the approach generalizes well on the unseen data. \cite{7159100} uses importance reweighting for learning in the presence of class conditional noise, and provide a method to estimate noise rates using density ratio estimation. 

Robust learning of ordinal regression models in the presence of label noise still remains an unaddressed problem.
In this paper, we propose an approach for learning robust ordinal regression in the presence of label noise. Our approach is inspired by the method of the unbiased estimator \cite{Natarajan:2013:LNL:2999611.2999745}. We have made the following contributions. 

\subsection*{Contributions}
\begin{enumerate}
\item We propose a label noise model for ordinal regression, namely inversely decaying noise. When the noise parameter is equal for all classes, we call it uniformly decaying noise. When the parameter changes with changing the class, we call it class conditional inversely decaying noise. 
\item We propose an unbiased estimator based approach for label noise robust ordinal regression. We work with losses ${l}_{CE}$ and ${l}_{IMC}$. We show that unbiased estimators $\Tilde{l}_{CE}$ and $\Tilde{l}_{IMC}$ are also rank consistent. 
\item We propose deep learning methods for robust ordinal regression which use $\Tilde{l}_{CE}$ and $\Tilde{l}_{IMC}$ as loss functions. We further show that stochastic gradient descent (SGD) on $\Tilde{l}_{CE}$ and $\Tilde{l}_{IMC}$ preserves the ordering of the thresholds and results in a rank consistent model.
\item We also provide generalization bounds for the proposed approach.
\item We experimentally show the effectiveness of the proposed approach on various datasets. We show that our approach can learn robust deep ordinal regression models well.
\end{enumerate}
To the best of our knowledge, this is the first attempt to address the label noise issue in ordinal regression.

\section{Ordinal Regression}
 Each example is of the form $(\x_i,y_i) \in \mathcal{X} \times \mathcal{Y} $, where $\mathcal{X} \subseteq \R^d$ and $\mathcal{Y} = \{1,\ldots,K\}$. The labels in $\mathcal{Y}$ are ordered, i.e. $1 \prec 2 \prec \ldots \prec K$.
 Let $\mathcal{D}$ be the unknown joint distribution on $\mathcal{X}\times \mathcal{Y}$ from which $N$ i.i.d. samples are drawn. 
 An ordinal regression function $f:\mathcal{X}\rightarrow \mathcal{Y}$ is described using a function $g:\mathcal{X}\rightarrow \mathbb{R}$ and thresholds $b_1,\ldots,b_K$ as follows.
\begin{align*}
f(\x) & = 1+\sum_{k=1}^{K-1} \mathbb{I}_{\{g(\x)+b_k>0\}}= \min_{i\in [K]} \{i: g(\x) + b_i \leq 0\}
\end{align*}
Let $\mathbf{b} = [b_1\;\ldots\;b_{K-1}]^T\in \mathbb{R}^{K-1}$. Thus, function $g(.)$ and thresholds $\mathbf{b}$ are the parameters to be optimized upon. We assume $b_K=-\infty$. We must ensure that $b_1 \geq \;\ldots \;\geq b_{K-1}$ to maintain the ordering among the classes \cite{Crammer:2001:PR:2980539.2980623,Li:2006:ORE:2976456.2976565}. 

\subsection{Loss Functions for Ordinal Regression}
We now describe commonly used loss functions which capture the discrepancy between the predicted label and the true label. 
\begin{enumerate}
    \item {\bf $l_{MAE}$: }Mean absolute error finds the absolute difference between the predicted label and the true label \cite{Antoniuk2016}. 
\begin{align}
\label{eq:MAE_Loss} &l_{MAE} (g(\x),\mathbf{b},y) = 
\sum_{i=1}^{y-1} \I_{\{g(\x) + b_i<0\}} + \sum_{i=y}^{K-1} \I_{\{g(\x) + b_i\geq 0\}}
\end{align}
$l_{MAE}=0$ whenever $-b_{y-1}\leq g(\x)\leq -b_{y}$. Optimizing $l_{MAE}$ is computationally hard as it is not continuous. Thus, in practice, we use convex surrogates of $l_{MAE}$ as loss functions to minimize the risk and find the parameters of $g(.)$ and thresholds $b_1,\ldots,b_{K-1}$.
\item {\bf $l_{IMC}$: }It is a
convex surrogate of $l_{MAE}$ \cite{Chu2005NewAT} which implicitly maintains the ordering of the thresholds $b_i$'s. 
\begin{align}
\nonumber &l_{IMC}(g(\x),\mathbf{b},y) = \sum_{i=1}^{y-1} \max\left(0,1-g(\x)-b_i\right)\\
& \;\;\;\;\;+\sum_{i=y}^{K-1}\max\left(0,1+g(\x)+b_i\right)\label{surrogate_loss}
\end{align}
For a given example-label pair $\{\x,y\}$, $l_{IMC}(g(\x),\mathbf{b},y)=0$ only when
$g(\x)+b_i \geq 1,  \forall i <y$ and $g(\x)+b_i \leq -1, \forall i \geq y$.
Let $z_i = \I_{\{i<y\}}-\I_{\{i\geq y\}},\;i\in[K]$. Thus, $z_i=1,\;\forall i<y$ and $z_i=-1,\;\forall i\geq y$. Thus, $l_{IMC}(g(\x),\mathbf{b},y) =0$ requires that $z_i(g(\x)+b_i)\geq 1,\;\forall i \in [K-1]$. Thus, 
\begin{eqnarray}
\nonumber l_{IMC}(f(\x),\mathbf{b},y) = \sum_{i=1}^{K-1} \max\left[0,1-z_i\left(g(\x)+b_i\right)\right]. \end{eqnarray}
In \cite{Chu2005NewAT}, it is shown that $l_{IMC}$ is implicitly rank consistent. Thus, at the optimal solution, $b_1 \leq \ldots \leq b_{K-1}$.
\item $l_{CE}$: Cross entropy loss \cite{DBLP:nn_based_approach,DBLP:journals/corr/abs-1901-07884} for ordinal regression is described as follows.
\begin{align*}
    l_{CE}(g(\x),\mathbf{b},y) &= - \sum_{j=1}^{K-1} [z_j \log(\sigma(g(\x)+b_j))  \\
    & + (1-z_j)\log (1- \sigma(g(\x)+b_j))] 
\end{align*}
where $\sigma(a)=(1+e^{-a})^{-1}$ is the sigmoid function. Also, $z_j=1,\;\forall j<y$ and $z_j=0,\;\forall j\geq y$. 
It is shown that $l_{CE}$ is rank consistent \cite{DBLP:journals/corr/abs-1901-07884}. Thus, minimizer of the risk under $l_{CE}$, will satisfy condition $b_1 \geq \ldots \geq b_{K-1}$.
\end{enumerate} 
 
\section{Label Noise Setting in Ordinal Regression}
Real-world datasets are seldom perfect and often suffer from various noise issues. One kind of noise in the data has noisy labels, where, we get corrupted samples $(\x_i,\tilde{y}_i),\;i=1\ldots N$, where $\tilde{y}_i$ are the noisy labels. The noisy label $\tilde{y}_i$ could be different from the true label $y_i$. A detailed discussion on the sources of noise can be found in \cite{noise_survey}. For classification problems, learning in presence of label noise is a well studied problem \cite{DBLP:noise_tolerance_nar,Natarajan:2013:LNL:2999611.2999745,ghosh2017robust,7159100}. 

Let $P(\tilde{y}=j|y=i,\x)=\eta_{(i,j)}(\x)$ be the probability of observing label $j$ for example $\x$ whose true label is $i$. Uniform label noise ($P(\tilde{y}=j|y=i,\x)=\eta,\;\forall i \neq j$ and $\forall \x$) and class conditional label noise ($P(\tilde{y}=j|y=i,\x)=\eta_{(i,j)},\;\forall \x \in C_i, i\neq j$) are some of the commonly used noise models \cite{noise_survey,ghosh2017robust}. 
For class conditional noise model, the noise model is entirely represented by noise matrix $\mathbf{N}$ such that $\mathbf{N}_{i,j}=\eta_{(i,j)}$.

\subsection{Label Noise Models for Ordinal Regression}
We note that the uniform and class conditional noise models described above do not take the ordinal aspect of the label into account due to the following reasoning.
In practice, when humans annotate the data (say rating a product), it is likely that even if there is an error in the labeling, the human has a sense of label ``category''. So they might be able to classify the product as good or bad, but there might be an error in imputing the correct rank. Thus, when they make errors in ranking, it is more likely that they choose neighboring ranks more often than the far away rank. Thus, it would make sense to study label noise models in which the noise probability of a label far away is less than that of a label nearer. 
With this in mind, we propose the following noise model. In the proposed noise model, the noise rate does not depend on $\x$. 
\begin{itemize}

\item {\bf Inversely decaying noise: }Here, the probability of mislabeling is inversely proportional to the absolute difference between the true rank and the rank of incorrect label. Thus, $\eta_{(i,j)} = \frac{\rho_i}{|i-j|},\;\forall i \neq j$ where $\rho_i$ is a parameter for class $i$. The diagonal element $\eta_{(i,i)}$ is defined as $\eta_{(i,i)} =  1 - \sum_{j=1,j \neq i}^K \eta_{(i,j)}$. If $\rho_i = \rho,\;\forall i$ then the noise model is called uniformly inversely decaying.  

{\bf Example 1: }Here, we see the noise matrix corresponding to a uniformly inversely decaying noise model. Let $\rho=0.15$, and there are 4 classes, then the noise matrix and its inverse are as follows.

\begin{table}[ht]
\centering
\resizebox{\columnwidth}{!}{
\begin{tabular}{cc} \
$\begin{bmatrix}
        0.725 & 0.15 & 0.075 & 0.05 \\
       0.15 & 0.625 & 0.15 & 0.075 \\
       0.075 & 0.15 & 0.625 & 0.15 \\
       0.05 & 0.075 & 0.15 & 0.725
\end{bmatrix}$ & $\begin{bmatrix}
        1.45 & -0.32 & -0.08 & -0.05 \\
       -0.32 & 1.77 & -0.36 & -0.09 \\
       -0.08 & -0.36 & 1.77 & -0.32 \\
       -0.05 & -0.09 & -0.32 & 1.45
\end{bmatrix}$ \\ 
$\mathbf{N}$ & $\mathbf{N}^{-1}$\end{tabular}}
\label{noise_matrix}
\end{table}

\end{itemize}

Observe that in the uniform version of the noise model, the probability of not flipping the label $\eta_{(i,i)}$ is maximum at the extremes and is minimum for mid-labels. Labels in the middle of the label range are more susceptible to noise as compared to labels at the end, which conforms to human behavior while ranking objects on an ordinal scale. Say a human is asked to rate a product on a scale of 1 to 10, 1 being poor quality and 10 being excellent. The human would be more confident when assigning extreme ratings, i.e., an excellent or terrible product is easy to distinguish. Thus, $\eta_{i,i}$ for extreme ratings would be high. On the other hand, the distinction between labels in the middle range is ambiguous. And hence identifying the true rating becomes more difficult in the middle range compared to the extremes. Hence, the decreasing values of $\eta_{i,i}$ in the middle range. 

\subsection{Properties of Noise Matrix of the Proposed Noise Models}
We observe the following properties of matrix $\mathbf{N}$. 
\begin{itemize}
    \item Since $ \eta_{(i,j)}$ is a function of $|i-j|$, matrix $ \mathbf{N}$ becomes symmetric.
    $\mathbf{N}^{-1}$ is also symmetric, because inverse of a symmetric matrix is symmetric.
    \item Each row (and column) has a sum of 1 as it represents a probability distribution of a random variable. $\sum_i \eta_{(i,j)} = \sum_j \eta_{(i,j)} = 1$. Thus, matrix $\mathbf{N}$ is doubly-stochastic. 
    \item If $\eta_{(i,i)} > 0.5$, then $\eta_{(i,i)} > \sum_{j, j \neq i} \eta_{(j,i)} $. This condition implies that the matrix $\mathbf{N}$ is (strictly) diagonally dominant. With this assumption, matrix $\mathbf{N}$ becomes non-singular \cite{Horn:2012:MA:2422911}.
    \item Row sum (and column sum) of $\mathbf{N}^{-1}$ is 1 as follows. Since $\mathbf{N}^{-1}\mathbf{N}=\mathbf{I}$, for all $j,k\in [K]$, we get,
    $\sum_{i=1}^K\eta_{(j,i)}\mathbf{N}^{-1}_{(i,k)} = \mathbb{I}_{\{j=k\}}$. Now, summing over $j=1\ldots K$, we get $\sum_{j=1}^K\sum_{i=1}^K\eta_{(j,i)}\mathbf{N}^{-1}_{(i,k)} = \sum_{j=1}^K\mathbb{I}_{\{j=k\}}$
    By rearranging the terms and using the fact that $\sum_{i=1}^K \eta_{(i,j)} = 1,\; \forall j$, we get,
    $\sum_{i=1}^K\mathbf{N}^{-1}_{(i,k)} = 1,\;\forall k$. Thus column sum of $\mathbf{N}^{-1}$ is $1$. Since $(\mathbf{N}^{-1})^T = \mathbf{N}^{-1}$, $\sum_{i=1}^K \mathbf{N}^{-1}_{(k,i)} = 1 $. Thus, row sum of $\mathbf{N}^{-1}$ is also $1$. The same can be seen in Example 1. 
    \item Every column (row) of $\mathbf{N}^{-1}$ has negative entries. This can verify it by contradiction. Suppose $ \mathbf{N}^{-1}$ has only non-negative elements in any column (all cannot be zero since $\mathbf{N}^{-1}$ is an invertible matrix). Consider the dot product of $i^{th}$ row of $\mathbf{N}$ and $j^{th}$ ($j\neq i$) column of $\mathbf{N}^{-1}$, which is  $\sum_{t=1}^K \eta_{(i,t)}\mathbf{N}^{-1}_{(t,j)}$. Since $\mathbf{N}^{-1}$ has all non-negative elements, we get, $\sum_{t=1}^K \eta_{(i,t)}\mathbf{N}^{-1}_{(t,j)}>0$. But, $\sum_{t=1}^K\eta_{(i,t)}\mathbf{N}^{-1}_{(t,j)} = 0,\forall j\neq i$, which is a contradiction. 
     Hence, in every column (and row) of $ \mathbf{N}^{-1}$, there is a negative element. The same can be seen in Example 1.  
\end{itemize}


\section{Robust Ordinal Regression in Presence of Label Noise}

In this section, we propose a methodology for robust ordinal regression. As discussed earlier, we get corrupted samples $(\x_i,\tilde{y}_i),\;i=1\ldots N$, where $\tilde{y}_i$ is the noisy label. Our approach is based on an unbiased estimator \cite{Natarajan:2013:LNL:2999611.2999745,Patrini_2017_CVPR}. Thus, we use unbiased estimator $\Tilde{l}(f(\x),\tilde{y})$ of the loss $l(f(\x),y)$. We use the noise matrix $N$ to construct the unbiased estimator $\Tilde{l}(f(\x),\tilde{y})$ of $l(f(\x),y)$, which means
\begin{equation} \label{eq:1}
    \mathbb{E}_{\Tilde{y}}[\Tilde{l}(f(\x),\tilde{y})] = l(f(\x),y).
\end{equation}
Thus, optimising the risk based on $\Tilde{l}(f(\x),\tilde{y})$ in presence of label noise results in optimising risk based on $l(f(\x),y)$ in the absence of noise.  
Using eq.(\ref{eq:1}), we get the following equation.
\begin{align}\label{sys:1}
     l(f(\x),y) = \mathbb{E}_{\tilde{y}}[\Tilde{l}(f(\x),\tilde{y})] = \sum_{\tilde{y}=1}^K \eta_{y,\tilde{y}} \Tilde{l}(f(\x),\tilde{y}) 
\end{align}
Let $\mathbf{\Tilde{L}} = \begin{bmatrix}
    \Tilde{l}(f(\x),1) & \Tilde{l}(f(\x),2)  & \dots  & \Tilde{l}(f(\x),K) 
\end{bmatrix}^T$ and
$\mathbf{L} = \begin{bmatrix}
    l(f(\x),1) & l(f(\x),2)  & \dots  & l(f(\x),K) 
\end{bmatrix}^T$, then the system of equations in (\ref{sys:1}) can be written as $\mathbf{N}\mathbf{\Tilde{L}}=\mathbf{L}$. 
Hence, we get $\mathbf{\Tilde{L}}=\mathbf{N}^{-1}\mathbf{L}$. Note that the transformation of $l$ to $\Tilde{l}$ depends only on the noise rates.
Also, function $\tilde{l}$ need not be convex even if we begin with convex $l$. 
In this paper, we work with losses $l_{CE}$ and $l_{IMC}$.
It can be easily verified that $\Tilde{l}_{CE}$ and $\Tilde{l}_{IMC}$ are no more convex functions.

\subsection{Rank Consistency of $\Tilde{l}_{CE}$ and $\Tilde{l}_{IMC}$}

The loss functions used in a robust method for ordinal regression also need to be rank consistent. While we know that both $l_{CE}$ and $l_{IMC}$ are rank consistent \cite{Chu2005NewAT,DBLP:journals/corr/abs-1901-07884}, it is required to show that $\Tilde{l}_{CE}$ and $\Tilde{l}_{IMC}$ are also  rank consistent. The following theorem proves the same.

\begin{theorem}\label{cons-ce}
$\Tilde{l}_{CE}$ and $\Tilde{l}_{IMC}$ are rank consistent.  
\end{theorem}
Proof of this Theorem is provided in the Supplementary file. We now discuss deep learning approach for learning robust ordinal regression models.

\subsection{Deep Learning Model for Robust Ordinal Regression}
In this paper, we propose deep neural network based approaches to ordinal regression using $\Tilde{l}_{CE}$ and $\Tilde{l}_{IMC}$ as loss functions. These approaches are robust to label noise.

\subsubsection{Approach 1: Based on Loss $\Tilde{l}_{CE}$} 
\begin{figure}[h]
\centering
\includegraphics[width=.9\linewidth, height=3.75cm]{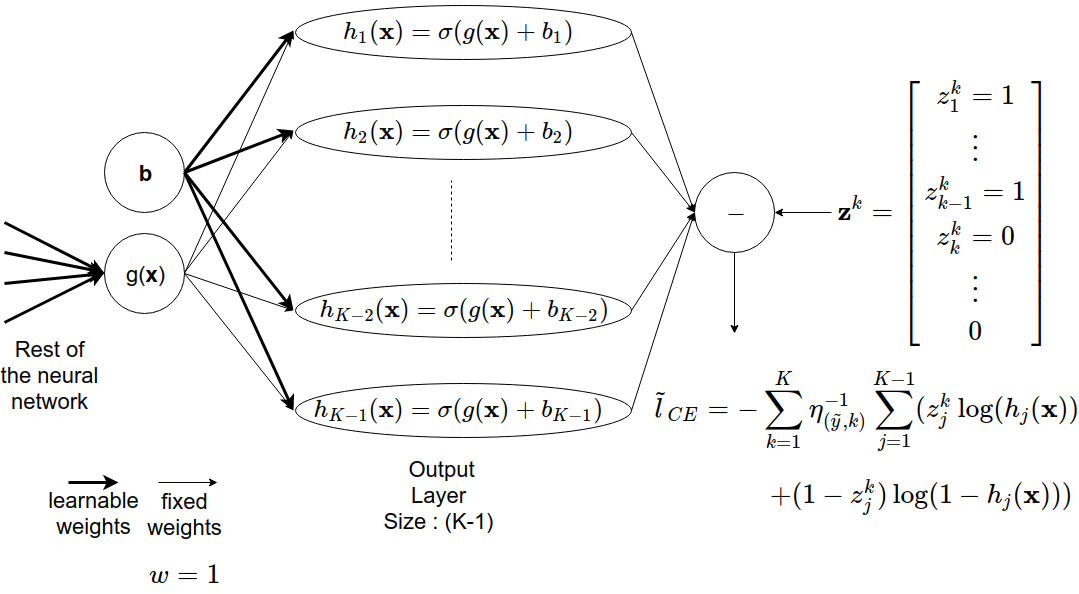} 
\caption{Neural network for robust ordinal regression based on $\Tilde{l}_{CE}$.}
\label{fig:subim1}
\end{figure}

We use the neural network architecture described in Figure~\ref{fig:subim1}.  The penultimate layer, whose output is denoted as $g(\x)$, shares a single weight  (but different bias) with all nodes in the pre-final layer. Let $h_j(\x) = \sigma(g(\x)+b_j)$, where $g(\x)$ is a function of input vector $\x$ computed using initial layers of the network. The pre-final layer in the network has $K-1$ nodes where $P(y>j|\x)=h_j(\x) = \sigma(g(\x)+b_j)$ is the output of $j^{th}$ node in that layer. $b_j$ is the bias term corresponding to the $j^{th}$ node in the pre-final layer. We use back-propagation algorithm (SGD) to minimize the loss function $\Tilde{l}_{CE}$ as follows.
\begin{align*}
    &\Tilde{l}_{CE}(g(\x),\mathbf{b},\Tilde{y}) = \sum_{j=1}^{K}\mathbf{N}^{-1}_{(\Tilde{y},j)}l_{CE}(g(\x),\mathbf{b}, j) \\  
    &= - \sum_{j=1}^{K}\mathbf{N}^{-1}_{(\Tilde{y},j)} \sum_{i=1}^{K-1} \left(\log h_i(\x)^{z_i^j}+\log(1-h_i(\x))^{(1-z_i^j)}\right)
\end{align*}
Where $\mathbf{N}^{-1}$ is the inverse of the noise matrix and $z_i^j=1,\;\forall i<j$ and $z_i^j=0,\;\forall i\geq j$.

We observe that the back-propagation algorithm for training the above network ensure the orderings among the thresholds in the expected sense as follows.
\begin{theorem}
SGD on $\Tilde{l}_{CE}$ maintains ordering among the thresholds. Let $b_i^t,\;i\in[K-1]$ be the thresholds at the $t^{th}$ round and $b_i^t-b_{i+1}^t\geq 0,\;i\in [K-1]$ holds true. Then, we observe that $\mathbb{E}_{\Tilde{y}^t}[b_i^{t+1}-b_{i+1}^{t+1}] \geq 0,\;\forall i \in [K-1]$.
\end{theorem}
The proof is given in the supplementary file. Note that the ordering consistency proof can be shown only in the expected sense because the back-propagation updates involve the terms containing $\mathbf{N}^{-1}_{(\Tilde{y},j)}$ which is a random variable. To normalize it, we need to take expectation with respect to $\Tilde{y}$. The theorem shows the correctness of the robust ordinal regression approach based on loss $\Tilde{l}_{CE}$. 

\subsubsection{Approach 2: based on Loss $\Tilde{l}_{IMC}$} 
\begin{figure}[h!]
\centering
\includegraphics[width=.9\linewidth, height=3.75cm]{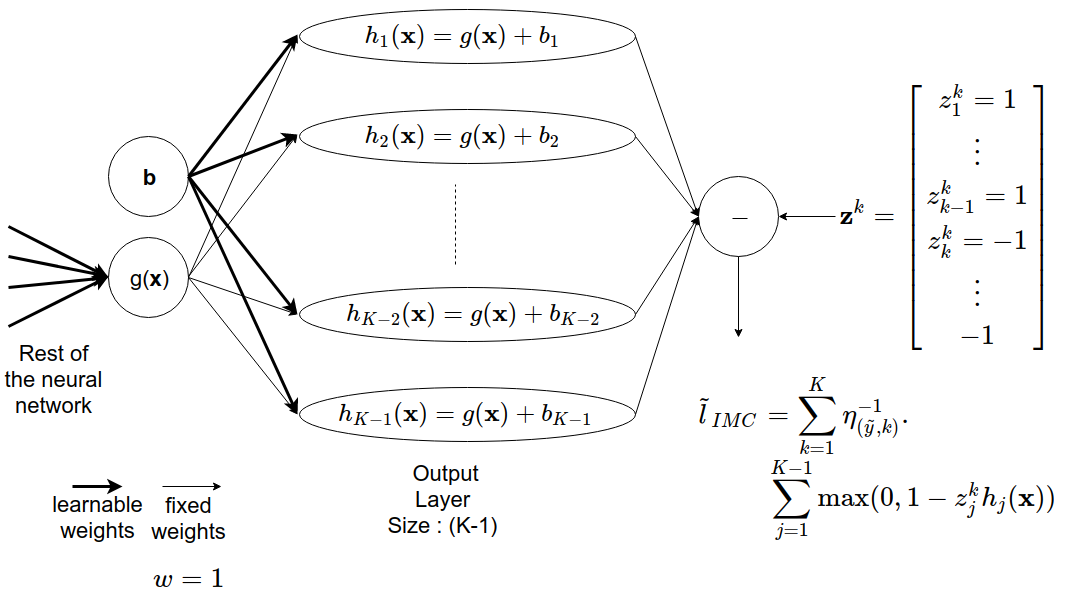} 
\caption{Neural network for robust ordinal regression using $\Tilde{l}_{IMC}$.}
\label{fig:subim2}
\end{figure}
We now give a neural network architecture for robust ordinal regression based on $\Tilde{l}_{IMC}$. The architecture is described in Figure~\ref{fig:subim2}. Similar to Approach~1, here also, the penultimate layer shares a single weight (but different bias) with all nodes in the pre-final layer. Pre-final layer has $K-1$ nodes whose outputs are $h_1(\x),\ldots,h_{K-1}(\x)$. Note that, here, $h_j(\x) = g(\x) + b_j$ where $g(\x) $ is some function of the weights of neural network leading to all but last layer and the input vector $\x_i$. We minimize the following loss function using back-propagation.
\begin{align*}
    &\Tilde{l}_{IMC}(g(\x^t),\mathbf{b},\Tilde{y}^t)=\sum_{j=1}^K \mathbf{N}^{-1}_{(\Tilde{y}^t,j)} \sum_{i=1}^{K-1} \left[1-z_i^j\left(g(\x^t)+b_i\right)\right]_+ 
\end{align*}
Where $z_i^j=1,\;\forall i<j$ and $z_i^j=-1,\;\forall i\geq j$.
\begin{theorem}
SGD on $\Tilde{l}_{IMC}$ maintains ordering among the thresholds. Let $b_i^t,\;i\in[K-1]$ be the thresholds at $t^{th}$ round and $b_i^t-b_{i+1}^t\geq 0,\;i\in [K-1]$ holds true. Then, we observe that $\mathbb{E}_{\Tilde{y}^t}[b_i^{t+1}-b_{i+1}^{t+1}] \geq 0,\;\forall i \in [K-1]$.
\end{theorem}
The proof is given in the supplementary file. Note that the ordering consistency proof can be shown only in the expected sense due to the similar reasons as Theorem 2.

%
\begin{table}[ht]
\begin{center}
\begin{tabular}{cc}
\includegraphics[scale=0.28]{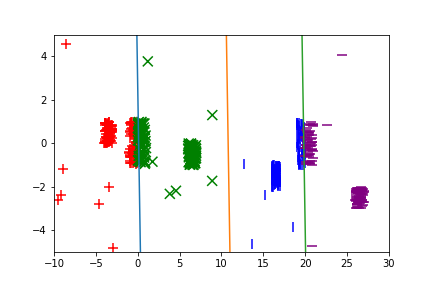} & \includegraphics[scale=.28]{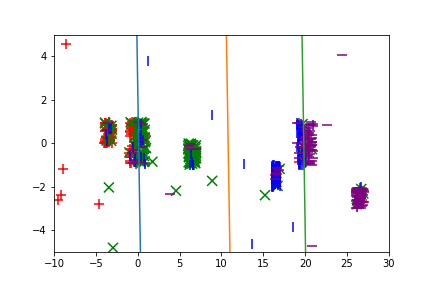} \\
a & b\\
\includegraphics[scale=.28]{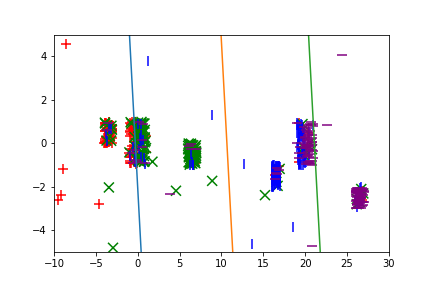}&\includegraphics[scale=.28]{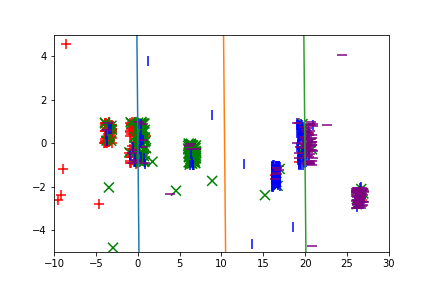}\\
 c & d \\
\end{tabular}
\caption{Results of different algorithms on Synthetic Dataset. (a) True classifier on clean data, (b) True classifier shown on noisy labels, (c) Classifier trained on noisy labels using $l$, (d) Classifier trained on noisy labels using $\tilde{l}$ with known noise rates }
\label{tab:gt}
 \end{center}
\end{table}

\begin{table*}[h!]
\footnotesize
\caption{}
\label{results}
\centering
\begin{tabular}{cccccc}
\hline 

\multicolumn{1}{c|}{} & \multicolumn{1}{c|}{\multirow{2}{*}{Loss fn}}   & \multicolumn{2}{c|}{Mean Absolute Error} & \multicolumn{2}{c}{Mean Zero-one Error} \\ \cline{3-6}

\multicolumn{1}{c|}{} & \multicolumn{1}{c|}{} & \multicolumn{1}{c|}{Clean Data}                               & \multicolumn{1}{c|}{Noisy data}    & \multicolumn{1}{c|}{Clean Data}    & \multicolumn{1}{c}{Noisy data} \\ \hline 

\multicolumn{1}{c|}{\multirow{7}{*}{Synth}}& \multicolumn{1}{c|}{$l_{CE}$} & \multicolumn{1}{c}{$0.03 \pm 0.00$}&
\multicolumn{1}{c|}{$0.20 \pm 0.02$}&
\multicolumn{1}{c}{$0.03 \pm 0.00$}&
\multicolumn{1}{c}{$0.20 \pm 0.02$}\\

\multicolumn{1}{c|}{}& \multicolumn{1}{c|}{$\tilde{l}_{CE}$-KR}   & \multicolumn{1}{c}{$0.03 \pm 0.00$}&
\multicolumn{1}{c|}{$\mathbf{0.06 \pm 0.01}$}&
\multicolumn{1}{c}{$0.03 \pm 0.00$}&
\multicolumn{1}{c}{$0.06 \pm 0.01$}\\

\multicolumn{1}{c|}{}& \multicolumn{1}{c|}{$\tilde{l}_{CE}$-EST}   & \multicolumn{1}{c}{$\mathbf{0.02 \pm 0.00}$}&
\multicolumn{1}{c|}{$0.12 \pm 0.01$}&
\multicolumn{1}{c}{$0.02 \pm 0.00$}&
\multicolumn{1}{c}{$0.12 \pm 0.01$}\\[4pt]

\multicolumn{1}{c|}{}& \multicolumn{1}{c|}{$l_{IMC}$}   & \multicolumn{1}{c}{$0.07 \pm 0.01$}&
\multicolumn{1}{c|}{$0.23 \pm 0.01$}&
\multicolumn{1}{c}{$0.07 \pm 0.01$}&
\multicolumn{1}{c}{$0.23 \pm 0.01$}\\

\multicolumn{1}{c|}{}& \multicolumn{1}{c|}{$\tilde{l}_{IMC}$-KR}   & \multicolumn{1}{c}{$0.07 \pm 0.01$}&
\multicolumn{1}{c|}{$\mathbf{0.12 \pm 0.05}$}&
\multicolumn{1}{c}{$0.07 \pm 0.01$}&
\multicolumn{1}{c}{$0.12 \pm 0.05$} \\

\multicolumn{1}{c|}{}& \multicolumn{1}{c|}{$\tilde{l}_{IMC}$-EST}   & \multicolumn{1}{c}{$\mathbf{0.03 \pm 0.04}$}&
\multicolumn{1}{c|}{$0.19 \pm 0.03$}&
\multicolumn{1}{c}{$0.03 \pm 0.04$}&
\multicolumn{1}{c}{$0.19 \pm 0.03$}\\ \hline


\multicolumn{1}{c|}{\multirow{7}{*}{Boston}}&
\multicolumn{1}{c|}{$l_{CE}$}   &\multicolumn{1}{c}{$\mathbf{0.36 \pm 0.05}$}&
\multicolumn{1}{c|}{$0.54 \pm 0.09$}&
\multicolumn{1}{c}{$0.33 \pm 0.04$}&
\multicolumn{1}{c}{$0.47 \pm 0.07$}\\

\multicolumn{1}{c|}{}& \multicolumn{1}{c|}{$\tilde{l}_{CE}$-KR}   & \multicolumn{1}{c}{$\mathbf{0.36 \pm 0.05}$}&
\multicolumn{1}{c|}{$\mathbf{0.52 \pm 0.05}$}&
\multicolumn{1}{c}{$0.33 \pm 0.04$}&
\multicolumn{1}{c}{$0.43 \pm 0.04$}\\

\multicolumn{1}{c|}{}& \multicolumn{1}{c|}{$\tilde{l}_{CE}$-EST}   & \multicolumn{1}{c}{$0.38 \pm 0.06$}&
\multicolumn{1}{c|}{$0.57 \pm 0.06$}&
\multicolumn{1}{c}{$0.34 \pm 0.05$}&
\multicolumn{1}{c}{$0.51 \pm 0.05$}\\ [4pt]

\multicolumn{1}{c|}{}& \multicolumn{1}{c|}{$l_{IMC}$}   & \multicolumn{1}{c}{$0.37 \pm 0.04$}&
\multicolumn{1}{c|}{$0.54 \pm 0.10$}&
\multicolumn{1}{c}{$0.34 \pm 0.03$}&
\multicolumn{1}{c}{$0.46 \pm 0.08$}\\

\multicolumn{1}{c|}{}& \multicolumn{1}{c|}{$\tilde{l}_{IMC}$-KR}   & \multicolumn{1}{c}{$0.37 \pm 0.04$}&
\multicolumn{1}{c|}{$\mathbf{0.50 \pm 0.05}$}&
\multicolumn{1}{c}{$0.34 \pm 0.03$}&
\multicolumn{1}{c}{$0.41 \pm 0.04$} \\

\multicolumn{1}{c|}{}& \multicolumn{1}{c|}{$\tilde{l}_{IMC}$-EST}   & \multicolumn{1}{c}{$0.37 \pm 0.05$}&
\multicolumn{1}{c|}{$0.53 \pm 0.08$}&
\multicolumn{1}{c}{$0.33 \pm 0.04$}&
\multicolumn{1}{c}{$0.47 \pm 0.06$}\\ \hline


\multicolumn{1}{c|}{\multirow{7}{*}{Abalone}}&  \multicolumn{1}{c|}{$l_{CE}$}   & \multicolumn{1}{c}{$\mathbf{0.42 \pm 0.02}$}&
\multicolumn{1}{c|}{$0.46 \pm 0.03$}&
\multicolumn{1}{c}{$0.40 \pm 0.02$}&
\multicolumn{1}{c}{$0.44 \pm 0.03$} \\

\multicolumn{1}{c|}{}& \multicolumn{1}{c|}{$\tilde{l}_{CE}$-KR}   & \multicolumn{1}{c}{$\mathbf{0.42 \pm 0.02}$}&
\multicolumn{1}{c|}{$\mathbf{0.44 \pm 0.02}$}&
\multicolumn{1}{c}{$0.40 \pm 0.02$}&
\multicolumn{1}{c}{$0.41 \pm 0.02$} \\

\multicolumn{1}{c|}{}& \multicolumn{1}{c|}{$\tilde{l}_{CE}$-EST}   & \multicolumn{1}{c}{$0.44 \pm 0.03$}&
\multicolumn{1}{c|}{$0.47 \pm 0.04$}&
\multicolumn{1}{c}{$0.42 \pm 0.03$}&
\multicolumn{1}{c}{$0.45 \pm 0.04$}\\ [4pt]

\multicolumn{1}{c|}{}& \multicolumn{1}{c|}{$l_{IMC}$}   & \multicolumn{1}{c}{$\mathbf{0.42 \pm 0.02}$}&
\multicolumn{1}{c|}{$0.54 \pm 0.03$}&
\multicolumn{1}{c}{$0.40 \pm 0.02$}&
\multicolumn{1}{c}{$0.50 \pm 0.02$} \\

\multicolumn{1}{c|}{}& \multicolumn{1}{c|}{$\tilde{l}_{IMC}$-KR}   & \multicolumn{1}{c}{$\mathbf{0.42 \pm 0.02}$}&
\multicolumn{1}{c|}{$\mathbf{0.43 \pm 0.02}$}&
\multicolumn{1}{c}{$0.40 \pm 0.02$}&
\multicolumn{1}{c}{$0.41 \pm 0.02$} \\

\multicolumn{1}{c|}{}& \multicolumn{1}{c|}{$\tilde{l}_{IMC}$-EST}   & \multicolumn{1}{c}{$0.45 \pm 0.02$}&
\multicolumn{1}{c|}{$0.48 \pm 0.02$}&
\multicolumn{1}{c}{$0.42 \pm 0.02$}&
\multicolumn{1}{c}{$0.45 \pm 0.01$}\\ \hline


\multicolumn{1}{c|}{\multirow{7}{*}{Computer}}&  \multicolumn{1}{c|}{$l_{CE}$}   & \multicolumn{1}{c}{$0.27 \pm 0.01$}&
\multicolumn{1}{c|}{$0.38 \pm 0.02$}&
\multicolumn{1}{c}{$0.26 \pm 0.01$}&
\multicolumn{1}{c}{$0.37 \pm 0.02$} \\

\multicolumn{1}{c|}{}& \multicolumn{1}{c|}{$\tilde{l}_{CE}$-KR}   & \multicolumn{1}{c}{$0.27 \pm 0.01$}&
\multicolumn{1}{c|}{$0.36 \pm 0.01$}&
\multicolumn{1}{c}{$0.26 \pm 0.01$}&
\multicolumn{1}{c}{$0.33 \pm 0.01$}\\

\multicolumn{1}{c|}{}& \multicolumn{1}{c|}{$\tilde{l}_{CE}$-EST}   & \multicolumn{1}{c}{$0.27 \pm 0.01$}&
\multicolumn{1}{c|}{$\mathbf{0.35 \pm 0.01}$}&
\multicolumn{1}{c}{$0.26 \pm 0.01$}&
\multicolumn{1}{c}{$0.33 \pm 0.01$}\\[4pt]

\multicolumn{1}{c|}{}& \multicolumn{1}{c|}{$l_{IMC}$}   & \multicolumn{1}{c}{$0.27 \pm 0.01$}&
\multicolumn{1}{c|}{$0.39 \pm 0.03$}&
\multicolumn{1}{c}{$0.26 \pm 0.01$}&
\multicolumn{1}{c}{$0.38 \pm 0.03$}\\

\multicolumn{1}{c|}{}& \multicolumn{1}{c|}{$\tilde{l}_{IMC}$-KR}   & \multicolumn{1}{c}{$0.27 \pm 0.01$}&
\multicolumn{1}{c|}{$0.35 \pm 0.02$}&
\multicolumn{1}{c}{$0.26 \pm 0.01$}&
\multicolumn{1}{c}{$0.33 \pm 0.02$}\\

\multicolumn{1}{c|}{}& \multicolumn{1}{c|}{$\tilde{l}_{IMC}$-EST}   & \multicolumn{1}{c}{$0.27 \pm 0.01$}&
\multicolumn{1}{c|}{$\mathbf{0.34 \pm 0.01}$}&
\multicolumn{1}{c}{$0.26 \pm 0.01$}&
\multicolumn{1}{c}{$0.32 \pm 0.01$}\\ \hline


\multicolumn{1}{c|}{\multirow{7}{*}{California}}&\multicolumn{1}{c|}{$l_{CE}$}   & \multicolumn{1}{c}{$0.30 \pm 0.01$}&
\multicolumn{1}{c|}{$0.38 \pm 0.02$}&
\multicolumn{1}{c}{$0.29 \pm 0.01$}&
\multicolumn{1}{c}{$0.33 \pm 0.01$} \\

\multicolumn{1}{c|}{}& \multicolumn{1}{c|}{$\tilde{l}_{CE}$-KR}   & \multicolumn{1}{c}{$0.30 \pm 0.01$}&
\multicolumn{1}{c|}{$\mathbf{0.34 \pm 0.01}$}&
\multicolumn{1}{c}{$0.29 \pm 0.01$}&
\multicolumn{1}{c}{$0.33 \pm 0.01$}\\ 

\multicolumn{1}{c|}{}&
\multicolumn{1}{c|}{$\tilde{l}_{CE}$-EST}   & \multicolumn{1}{c}{$0.30 \pm 0.01$}&
\multicolumn{1}{c|}{$0.35 \pm 0.01$}&
\multicolumn{1}{c}{$0.29 \pm 0.01$}&
\multicolumn{1}{c}{$0.34 \pm 0.01$}\\[4pt]

\multicolumn{1}{c|}{}& \multicolumn{1}{c|}{$l_{IMC}$}   & \multicolumn{1}{c}{$0.31 \pm 0.01$}&
\multicolumn{1}{c|}{$0.41 \pm 0.04$}&
\multicolumn{1}{c}{$0.30 \pm 0.01$}&
\multicolumn{1}{c}{$0.41 \pm 0.04$}\\

\multicolumn{1}{c|}{}& \multicolumn{1}{c|}{$\tilde{l}_{IMC}$-KR}   & \multicolumn{1}{c}{$0.31 \pm 0.01$}&
\multicolumn{1}{c|}{$\mathbf{0.34 \pm 0.01}$}&
\multicolumn{1}{c}{$0.30 \pm 0.01$}&
\multicolumn{1}{c}{$0.33 \pm 0.01$}\\

\multicolumn{1}{c|}{}& \multicolumn{1}{c|}{$\tilde{l}_{IMC}$-EST}   & \multicolumn{1}{c}{$0.31 \pm 0.00$}&
\multicolumn{1}{c|}{$0.35 \pm 0.01$}&
\multicolumn{1}{c}{$0.30 \pm 0.01$}&
\multicolumn{1}{c}{$0.34 \pm 0.01$}\\ \hline


\multicolumn{1}{c|}{\multirow{7}{*}{MSLR}}&  \multicolumn{1}{c|}{$l_{CE}$}   & \multicolumn{1}{c}{$0.55 \pm 0.01$}&
\multicolumn{1}{c|}{$0.63 \pm 0.01$}&
\multicolumn{1}{c}{$0.49 \pm 0.01$}&
\multicolumn{1}{c}{$0.59 \pm 0.01$} \\

\multicolumn{1}{c|}{}& \multicolumn{1}{c|}{$\tilde{l}_{CE}$-KR}   & \multicolumn{1}{c}{$0.55 \pm 0.01$}&
\multicolumn{1}{c|}{$\mathbf{0.55 \pm 0.01}$}&
\multicolumn{1}{c}{$0.49 \pm 0.01$}&
\multicolumn{1}{c}{$0.49 \pm 0.02$} \\

\multicolumn{1}{c|}{}& \multicolumn{1}{c|}{$\tilde{l}_{CE}$-EST}   & \multicolumn{1}{c}{$\mathbf{0.53 \pm 0.01}$}&
\multicolumn{1}{c|}{$0.62 \pm 0.01$}&
\multicolumn{1}{c}{$0.46 \pm 0.01$}&
\multicolumn{1}{c}{$0.52 \pm 0.02$} \\[4pt]

\multicolumn{1}{c|}{}& \multicolumn{1}{c|}{$l_{IMC}$}   & \multicolumn{1}{c}{$\mathbf{0.55 \pm 0.01}$}&
\multicolumn{1}{c|}{$0.71 \pm 0.01$}&
\multicolumn{1}{c}{$0.49 \pm 0.01$}&
\multicolumn{1}{c}{$0.68 \pm 0.01$} \\

\multicolumn{1}{c|}{}& \multicolumn{1}{c|}{$\tilde{l}_{IMC}$-KR}   & \multicolumn{1}{c}{$\mathbf{0.55 \pm 0.01}$}&
\multicolumn{1}{c|}{$\mathbf{0.55 \pm 0.01}$}&
\multicolumn{1}{c}{$0.49 \pm 0.01$}&
\multicolumn{1}{c}{$0.50 \pm 0.01$} \\

\multicolumn{1}{c|}{}& \multicolumn{1}{c|}{$\tilde{l}_{IMC}$-EST}   & \multicolumn{1}{c}{$0.56 \pm 0.01$}&
\multicolumn{1}{c|}{$0.66 \pm 0.01$}&
\multicolumn{1}{c}{$0.49 \pm 0.01$}&
\multicolumn{1}{c}{$0.55 \pm 0.01$} \\ \hline

\end{tabular}
\end{table*}

\begin{table}[h!] 
\footnotesize
\caption{Iterations with unordered thresholds / total iterations}
\centering
\label{thresh}
\begin{tabular}{cccc} 
\hline 
\multicolumn{1}{c|}{} & \multicolumn{1}{c|}{\multirow{2}{*}{Loss fn}} & \multicolumn{1}{c|}{\multirow{2}{*}{Clean Data}}   & \multicolumn{1}{c}{\multirow{2}{*}{Noisy Data}}  \\ 

\multicolumn{1}{c|}{} & \multicolumn{1}{c|}{} & \multicolumn{1}{c|}{}& \multicolumn{1}{c}{}\\ \hline 

\multicolumn{1}{c|}{\multirow{2}{*}{Synth}}&  \multicolumn{1}{c|}{$l_{CE}-EST$} &
\multicolumn{1}{c}{$0.4 / 67200$}&
\multicolumn{1}{c}{$0.7 / 67200$}\\
\multicolumn{1}{c|}{}& \multicolumn{1}{c|}{$l_{IMC}-EST$}   &\multicolumn{1}{c}{$0.0 / 67200$}&
\multicolumn{1}{c}{$0.2 / 67200$} \\ \hline

\multicolumn{1}{c|}{\multirow{2}{*}{California}}&  \multicolumn{1}{c|}{$l_{CE}-EST$} &
\multicolumn{1}{c}{$0.0 / 247680$}&
\multicolumn{1}{c}{$0.2 / 247680$}\\
\multicolumn{1}{c|}{}& \multicolumn{1}{c|}{$l_{IMC}-EST$}   &\multicolumn{1}{c}{$0.2 / 247680$}&
\multicolumn{1}{c}{$0.2 / 247680$} \\ \hline

\end{tabular}
\end{table}

\subsection{Estimating Noise Rates}

We use the noise rate estimation method proposed in \cite{Patrini_2017_CVPR}   for our problem. To estimate the noise rates, we treat ordinal regression as a multiclass problem. We could not come up with an approach to estimate noise rates tailored to ordinal regression. The reason behind is as follows. The approach proposed in \cite{Patrini_2017_CVPR} requires the datasets to have a perfect representative class label, which is often not satisfied in ordinal regression problems. Nevertheless, we found the approach to work well for our use case.

\subsection{Generalization Bounds}
We represent the total risk of loss functions $l_{IMC}$ and $l_{CE}$ as sum of risks of $K-1$ binary classifiers i.e \begin{align*} & R_{l,\mathit{D}}(g,\mathbf{b}) = \mathbb{E}_{\mathit{D}}[l(g(\x),\mathbf{b},y)] \\ & = \sum_{i=1}^{K-1} \mathbb{E}_{\mathit{D}}[l^i(g(\x),\mathbf{b},z_i^{y})] = \sum_{i=1}^{K-1} R_{l^i,\mathit{D}}(g,\mathbf{b}))
\end{align*}
where $l^i$ , $1\leq i \leq K-1$ represents the loss at the $i^{th}$ binary classifier. Let $\hat{f} \in \arg \min_{f \in \mathcal{F}} \hat{R}_{\tilde{l},S}(f) $ and 
$f^{*} \in \arg \min_{f \in \mathcal{F}} {R}_{l,\mathit{D}}(f)$ where $\mathcal{F}$ is the hypothesis class of the function $f$. This proof is inspired from Theorem 3 in \cite{Natarajan:2013:LNL:2999611.2999745}. 

\begin{theorem} \label{GB}
If $\mathfrak{R}(\mathcal{F})$ is the Rademacher complexity of the function class $\mathcal{F}$, and the loss l is L-Lipschitz, then with probability atleast $1-\delta$, 
\begin{align*}R_{l,\mathit{D}}(\hat{f})  \leq R_{l,\mathit{D}}(f^{*}) + 2(K-1) \Big(2ML_{\rho} \mathfrak{R}(\mathcal{F}) + \sqrt{\frac{log(1/\delta)}{2n}}\Big)
\end{align*}
\end{theorem}

The proof is available in the supplementary file. The Theorem~\ref{GB} shows that the  risk (on clean distribution $\mathit{D}$) of classifier $\hat{f}$ learnt under $\tilde{l}$ with label noise  is bounded by risk of classifier from $l$ without label noise. In using an unbiased estimator, we pay the price of bigger Lipschitz-constant for $\tilde{l}$ and thus needs a larger training sample to generalize well. 

\section{Experiments} \label{experiments}
We conduct experiments on synthetic and real datasets \footnote{The California Housing dataset can be found \href{}{http://lib.stat.cmu.edu/datasets/}} \footnote{The other datasets can be found at  \href{}{https://www.dcc.fc.up.pt/~ltorgo/Regression/DataSets.html}}  to illustrate the effectiveness of our approach. Each feature is scaled to have 0 mean and unit variance coordinate wise. For hyperparameter tuning, we make a grid for two parameters: learning rate, size of hidden layer - and use  5-fold cross-validation to select the optimal parameters for the model. The number of epochs is chosen to be 300, and we observe that the loss converges for all models. We chose  AdamW optimiser  as the optimising algorithm with $\beta_1, \beta_2$=(0.9,0.999) and $L_2$ penalty with weight decay of $0.01$ - the default parameters \cite{article_Adamw}. We use ReLu as the activation function in the hidden layers for all datasets except for synthetic, where we use a Linear function to be able to demonstrate our method pictographically.  All codes are written in PyTorch. The hyperparameters are tuned for noisy labels (both training and testing) using loss $l$,  and the same parameters are used for the other two models, as described below. This is to ensure a strict test for the performance of the proposed unbiased estimator.     

We generate noisy labels with uniform inversely decaying model using $\rho =0.15 $. 

\subsection*{Estimating noise rates} We train multiclass neural network with negative log likelihood loss. Following \cite{Patrini_2017_CVPR}, instead of taking $\arg \max$ to chose $\x_i$ we use $99$ percentile. The noise matrix is constructed, and sample constructed noise matrices can be found in the supplementary file.  

We split the dataset into 80\% and 20\% independently 20 times, and train the following models corresponding to both $l_{CE}$ and $l_{IMC}$. (1) $l$:trained using loss function $l$; (2) $\tilde{l}$-KR: trained using $\tilde{l}$ with known noise rates; (3) $\tilde{l}$-EST: trained using $\tilde{l}$ with estimated noise rates.
    
The mean of $MAE$ and $0-1$ error along with the standard deviation of these 20 trials are presented in table \ref{results}.

%

\subsection{Discussion}
We compare our method of unbiased estimator with the benchmark deep learning ordinal regression method of using $l$ \cite{DBLP:nn_based_approach,DBLP:journals/corr/abs-1901-07884}. The performance of $l_{CE}$ is consistently better than $l_{IMC}$. The performance of $l$ with noise is seen to degrade more for $l_{IMC}$ compared to $l_{CE}$. As seen in the Table.~\ref{tab:gt} we see that the noise changes the orientation of the classifier when trained using $l$, giving sub-optimal results. The $\tilde{l}$ accounts for the noise and gives a robust classifier. 

We also observe that the noise rate estimates were good to work in the unbiased estimator where the datasets had $MAE$ error of less than $35\%$ on clean data (Synth, Computer, California). Here the performance of $\tilde{l}$-KR and $\tilde{l}$-EST were at par. The large deviations in noise rates estimates comes because of violations of Statement-1 in Theorem-3 \cite{Patrini_2017_CVPR}, which is more likely when the $MAE$ is high even for clean data. We also observe that the unbiased estimator performs well even with approximate noise rates. This is in line with observations made by \cite{Natarajan:2013:LNL:2999611.2999745}. 

The performance of $\tilde{l}$ for Boston data is just at par with $l$ because the Boston dataset is small ($\approx$ 500 samples). For very large dataset MSLR, $\tilde{l}$-KR with noise performs as good as $l$ on clean data.  This indicates that a comparatively larger number of samples are needed for the unbiased estimator to be able to perform well and be noise-robust. This can also be seen from the generalization error term in Theorem~\ref{GB}. 

\subsection*{Rank consistency} The proofs of rank consistency use expectation in the difference between adjacent thresholds. To check for rank consistency, we check the ordering of threshold after each update to the neural network. If the thresholds aren't ordered, we flag the iteration. To save space, we only report the average of the number of iterations with unordered thresholds for $\tilde{l}_{CE}$-EST and $\tilde{l}_{IMC}$-EST for synthetic and California housing datasets in  Table \ref{thresh} over the 20 iterations. The results for other datasets are similar.  We observe that even if thresholds are reversed for some iteration, they get corrected quickly. All the final models were rank consistent.


\section{Conclusions and Future Work}
In this paper, we propose a label noise model for ordinal regression. We then propose an unbiased estimator approach for learning robust ordinal regression models. We show that the models under $\tilde{l}_{CE}$ and $\tilde{l}_{IMC}$ are also rank consistent, which is a desirable property for ordinal regression. We empirically verify the efficiency of the proposed end-to-end method on synthetic as well as real datasets. While performing the experiments for $\tilde{l}$-EST, we do not make any assumptions on the noise model.

For further study, we could consider coming up with methods to make  estimating noise rates more reliable. We could also look into the effects of non-symmetric label noise on the model. This is the first study of ordinal regression under label noise.

\appendix
\section{Proof of Theorem~1}
\subsection{Rank consistency proof for $\Tilde{l}_{CE}$}
We need to show that $b_1 \geq b_2 \geq \ldots \geq b_{K-1}$ at the optimal solution. 
Let $\mathbf{b} = [b_1, b_2,..,b_{K-1}]^T  $, and $\mathbf{b}^*$ be the optimal value of $\mathbf{b}$. 
Let $(\x_i,\Tilde{y}_i),\;i=1\ldots N$ be the training set.
Let for some $j$ suppose $b_j < b_{j+1}$. Then we show that by replacing $b_j$ with $b_{j+1}$ or replacing $b_{j+1}$ with $b_j$ can further decrease the loss $\mathbf{\Tilde{L}_{CE}} = \mathbf{N}^{-1}\mathbf{L}_{CE}$, where $\mathbf{\Tilde{L}_{CE}}=\begin{bmatrix}
    \Tilde{l}_{CE}(g(\x),\mathbf{b},1) \\ \vdots  \\ \Tilde{l}_{CE}(g(\x),\mathbf{b},j+1)  \\ \vdots  \\ \Tilde{l}_{CE}(g(\x),\mathbf{b},K) 
\end{bmatrix}$ and $\mathbf{L}_{CE}=\begin{bmatrix}
    l_{CE}(g(\x),\mathbf{b},1) \\ \vdots  \\l_{CE}(g(\x),\mathbf{b},j+1)  \\ \vdots  \\ l_{CE}(g(\x),\mathbf{b},K) 
\end{bmatrix}$. We see that the change in $\mathbf{\Tilde{L}_{CE}}$ depends on $\mathbf{L}_{CE}$ as follows.
\begin{align*}
&\Delta \mathbf{\Tilde{L}_{CE}} = \mathbf{N^{-1}} \Delta \mathbf{ L}_{CE} \\ 
      \Rightarrow & \Delta \mathbf{\Tilde{L}_{CE}} = \mathbf{N}^{-1}   \begin{bmatrix}
    \Delta l_{CE}(g(\x),\mathbf{b},1)   \\ \vdots\\ \Delta l_{CE}(g(\x),\mathbf{b},j+1)  \\ \vdots  \\ \Delta l_{CE}(g(\x),\mathbf{b},K) 
\end{bmatrix}
\end{align*}
We now have to find the change $\Delta l_{CE}(g(\x_i),\mathbf{b},k)$ for every $i\in [N]$ and every $k\in [K-1]$. In order to do that, we first
consider the following three partitions of the training set.
\begin{align*}
    A_1 &= \{\x_i : y_i < j+1 \implies z_{y_i}^j = z_{y_i}^{j+1} = 0 \} \\
    A_2 &= \{\x_i : y_i > j+1 \implies z_{y_i}^j = z_{y_i}^{j+1} = 1 \} \\
    A_3 &= \{\x_i : y_i = j+1 \implies z_{y_i}^j = 1,  z_{y_i}^{j+1} = 0 \}
\end{align*}
The above three sets are mutually exclusive and exhaustive, i.e., $A_1 \cup A_2 \cup A_3 = \{\x_1,\ldots,\x_N\}$.
Let $h_j(\x) = \sigma(g(\x)+b_j)$. Now, we first find the change $\Delta l_{CE}(g(\x_i),\mathbf{b},k)$ for every $k\in [K-1]$ in these sets individually.
\begin{enumerate}
    \item {\bf Change in $l_{CE}$ for $\x_i \in A_1$: } The change in $l_{CE}$ when replacing $b_j$ with $b_{j+1}$ is, $$\Delta^a l_{CE}(g(\x_i),\mathbf{b},y_i) = \log (1 - h_{j}(\x_i)) -\log(1-h_{j+1}(\x_i)).$$
The change in $l_{CE}$ when replacing  $b_{j+1}$ with $b_j$ is,
$$ \Delta^b l_{CE}(g(\x_i),\mathbf{b},y_i) = \log (1 - h_{j+1}(\x_i)) - \log(1-h_{j}(\x_i)) .$$
The total change in loss $l_{CE}$ after swapping $b_j$ and $b_{j+1}$ is   $\Delta l_{CE}(g(\x),\mathbf{b},y_i)=(\Delta^a  + \Delta^b )l_{CE}(g(\x),\mathbf{b},y_i) = 0$.
\item {\bf Change in $l_{CE}$ for $A_2$: } 
The change in $l_{CE}$ when replacing $b_j$ with $b_{j+1}$ is 
$$\Delta^a l_{CE}(g(\x),\mathbf{b},y_i) = \log(h_{j}(\x))- \log ( h_{j+1}(\x)).$$
The change in $l_{CE}$ replacing  $b_{j+1}$ with $b_j$ 
$$ \Delta^b l_{CE}(g(\x),\mathbf{b},y_i) = \log(h_{j+1}(\x))-  \log(h_{j}(\x)) .$$
The total change in loss $L_{CE}$ after swapping $b_j$ and $b_{j+1}$ is   $(\Delta^a  + \Delta^b) l_{CE}(g(\x),\mathbf{b},y_i) = 0$.
\item {\bf Change in $l_{CE}$ for $A_3$: }
The change in $l_{CE}$ when replacing $b_j$ with $b_{j+1}$ is 
$$\Delta^a l_{CE}(g(\x),\mathbf{b},y_i) = \log(h_{j}(\x))- \log (h_{j+1}(\x)).$$
The change in $l_{CE}$ replacing  $b_{j+1}$ with $b_j$ 
$$ \Delta^b l_{CE}(g(\x),\mathbf{b},y_i) = - \log (1- h_{j}(\x)) - \log(1-h_{j+1}(\x)).$$
The total change in loss $l_{CE}$ after swapping $b_j$ and $b_{j+1}$ and given that $b_j \geq b_{j+1}$ is   
\begin{align*}
    (\Delta^a  + & \Delta^b) l_{CE} (g(\x),\mathbf{b},y_i) = \log(h_{j}(\x))-\log ( h_{j+1}(\x))  \\
    & \quad -   (\log(1-h_{j+1}(\x)) - \log (1- h_{j}(\x)) \\
    & < 0
\end{align*}          
 \end{enumerate} 
Hence  
\begin{equation*}
    (\Delta^a  + \Delta^b) l_{CE}(g(\x),\mathbf{b},y_i) = \begin{cases}
      \delta, & \text{if}\ y_i=j+1 \\
      0, & \text{if}\ y_i \neq j+1
    \end{cases}
\end{equation*}
for some $\delta<0$. Now consider the equations 
\begin{align*}
      \Rightarrow &(\Delta^a  + \Delta^b) \mathbf{\Tilde{L}_{CE}} = \mathbf{N}^{-1}   \begin{bmatrix}
    (\Delta^a  + \Delta^b)l_{CE}(g(\x),\mathbf{b},1)   \\ \vdots\\ (\Delta^a  + \Delta^b) l_{CE}(g(\x),\mathbf{b},j+1)  \\ \vdots  \\ (\Delta^a  + \Delta^b)l_{CE}(g(\x),\mathbf{b},K) 
\end{bmatrix}\\
\Rightarrow &\begin{bmatrix}
    (\Delta^a  + \Delta^b)\Tilde{l}_{CE}(g(\x),\mathbf{b},1) \\  \vdots  \\ (\Delta^a  + \Delta^b)\Tilde{l}_{CE}(g(\x),\mathbf{b},j+1)  \\ \vdots  \\ (\Delta^a  + \Delta^b)\Tilde{l}_{CE}(g(\x),\mathbf{b},K) \end{bmatrix} = \mathbf{N^{-1}}\begin{bmatrix}
    0 \\ \vdots\\    \delta \\ \vdots \\ 0\end{bmatrix}
\end{align*}
The change in loss $\Tilde{l}_{CE}$ is as follows.

\begin{align*}
 (\Delta^a  + \Delta^b) {R}_{\rho} &=  (\Delta^a  + \Delta^b)\mathbb{E}_{\tilde{y}} [\Tilde{l}_{CE}(g(\x),\mathbf{b},\Tilde{y})]\\
   &= \mathbb{E}_{\tilde{y}} [(\Delta^a  + \Delta^b)\Tilde{l}_{CE}(g(\x),\mathbf{b},\Tilde{y})]\\
    &= \mathbb{E}_{\tilde{y}} [\mathbf{N}^{-1}_{(\tilde{y},i+1)} \delta]\\ 
    &= \delta \mathbb{E}_{\tilde{y}} [\mathbf{N}^{-1}_{(\tilde{y},i+1)} ]\\ 
 &=\delta \sum_{k=1}^K P(\Tilde{y}=k) \mathbf{N}^{-1}_{(k,i+1)}\\
  &= \delta \sum_{k=1}^K \mathbf{N}^{-1}_{(k,i+1)} \sum_{j=1}^K P(y=j)P(\Tilde{y}=k|y=j)  \\
  &=\delta \sum_{j=1}^K P(y=j)\sum_{k=1}^K  \eta_{(j,k)} \mathbf{N}^{-1}_{(k,i+1)}\\
  &=\delta \sum_{j=1}^K P(y=j)\mathbb{I}_{\{j=i+1\}} \\
  &=\delta P(y=i+1)\leq 0 
 \end{align*}

That means by swapping $b_j$ and $b_{j+1}$, we can further reduce the total loss $\Tilde{L}_{CE}$, which is a contradiction to the assumption that $\mathbf{b}$ is the optimal solution under  $\Tilde{L}_{CE}$.  This completes the proof that $\tilde{l}_{CE}$ is also rank consistent. 

\subsection{Rank consistency proof for $\Tilde{l}_{IMC}$}
We need to show that $b_1 \geq b_2 \geq \ldots \geq b_{K-1}$ at the optimal solution. We use a similar methodology as Theorem 1 Section 1.1 to prove this. Let $\mathbf{b} = [b_1, b_2,..,b_{K-1}]^T  $, and $\mathbf{b}^*$ be the optimal value of $\mathbf{b}$. 

Let for some $j$ suppose $b_j < b_{j+1}$. Then we show that by replacing $b_j$ with $b_{j+1}$ or replacing $b_{j+1}$ with $b_j$ can further decrease the loss $\mathbf{\Tilde{L}} = \mathbf{N}^{-1}\mathbf{L}$.  
Consider the following sets.
\begin{align*}
    A_1 &= \{i : y_i < j+1 \implies z_{y_i}^j = z_{y_i}^{j+1} = -1 \} \\
    A_2 &= \{i : y_i > j+1 \implies z_{y_i}^j = z_{y_i}^{j+1} = +1 \} \\
    A_3 &= \{i : y_i = j+1 \implies z_{y_i}^j = -1,  z_{y_i}^{j+1} = +1 \}
\end{align*}
The above three sets are mutually exclusive and exhaustive, i.e., $A_1 \cup A_2 \cup A_3 = \{1,2,..,N\}$. 
\begin{enumerate}
    \item {\bf Change in $l_{IMC}$ for $A_1$: } The change in $l_{IMC}$ when replacing $b_j$ with $b_{j+1}$ is \begin{align*}
        \Delta^a l_{IMC}(f(\x),y_i) &= \max(0, -1(g(\x_i)+b_{j+1})+1)  \\ & \quad - \max(0, -1(g(\x_i)+b_{j})+1)
    \end{align*} 
The change in $l_{IMC}$ when replacing  $b_{j+1}$ with $b_j$ 
\begin{align*} \Delta^b l_{IMC}(f(\x),y_i) &= \max(0, -1(g(\x_i)+b_{j})+1)  \\ & \quad - \max(0, -1(g(\x_i)+b_{j+1})+1) \end{align*}
The total change in loss $L_{IMC}$ after swapping $b_j$ and $b_{j+1}$ is   $(\Delta^a  + \Delta^b )l_{IMC}(f(\x),y_i) = 0$
\item {\bf Change in $l_{IMC}$ for $A_2$: } 
The change in $l_{IMC}$ when replacing $b_j$ with $b_{j+1}$ is 
\begin{align*} \Delta^a l_{IMC}(f(\x),y_i) &=\max(0, +1(g(\x_i)+b_{j+1})+1) \\ & \quad - \max(0, +1(g(\x_i)+b_{j})+1)\end{align*}
The change in $l_{IMC}$ replacing  $b_{j+1}$ with $b_j$ 
\begin{align*}  \Delta^b l_{IMC}(f(\x),y_i) &= \max(0, +1(g(\x_i)+b_{j})+1) \\ & \quad - \max(0, +1(g(\x_i)+b_{j+1})+1)\end{align*}
The total change in loss $L_{IMC}$ after swapping $b_j$ and $b_{j+1}$ is   $(\Delta^a  + \Delta^b) l_{IMC}(f(\x),y_i) = 0$
\item {\bf Change in $l_{IMC}$ for $A_3$: }
The change in $l_{IMC}$ when replacing $b_j$ with $b_{j+1}$ is 
\begin{align*}
\Delta^a l_{IMC}(f(\x),y_i) &= \max(0, -1(g(\x_i)+b_{j+1})+1) \\ & \quad - \max(0, -1(g(\x_i)+b_{j})+1) \\
                            &= \max(0,- b_{j+1}-g(\x_i)+1) \\ & \quad - \max(0, -b_{j} -g(\x_i)+1+1) \\
                            & \leq 0 
\end{align*}
The change in $l_{IMC}$ replacing  $b_{j+1}$ with $b_j$ 
\begin{align*}
    \Delta^b l_{IMC}(f(\x),y_i) &= \max(0, +1(g(\x_i)+b_{j})+1) \\ & \quad - \max(0, +1(g(\x_i)+b_{j+1})+1)\\
                            &= \max(0, g(\x_i)+b_{j}+1) \\ & \quad - \max(0, g(\x_i)+b_{j+1}+1) \\
                            & \leq 0 
\end{align*}

Now suppose $\Delta^a l_{IMC}(f(\x),y_i)= 0$. Since $b_j < b_{j+1}$ we have 
\begin{align*}
    g(\x_i) + b_j &\geq 1  \\
\& \qquad   g(\x_i) + b_{j+1} &> 1    \numberthis \label{dela}
\end{align*}

From \ref{dela}, we have in $\Delta^b l_{IMC}(f(\x),y_i)$,
\begin{align*}
    \Delta^b l_{IMC}(f(\x),y_i)  &= \max(0, g(\x_i)+b_{j}+1) \\ & \quad - \max(0, g(\x_i)+b_{j+1}+1) \\
                            &= b_{j+1} - b_j \\
                            & < 0
\end{align*}

Similarly, if $\Delta^b l_{IMC}(f(\x),y_i) = 0 $, we will have $\Delta^a l_{IMC}(f(\x),y_i) < 0$

The total change in loss $l_{IMC}$ after swapping $b_j$ and $b_{j+1}$ and given that $b_j < b_{j+1}$ is   
\begin{align*}
    (\Delta^a  + \Delta^b) l_{IMC} (f(\x),y_i) & < 0
\end{align*}          
 \end{enumerate} 
Hence  
\begin{equation*}
    (\Delta^a  + \Delta^b) l_{IMC}(f(\x),y_i) = \begin{cases}
      \delta, & \text{if}\ y_i=j+1 \\
      0, & \text{if}\ y_i \neq j+1
    \end{cases}
\end{equation*}
for some $\delta<0$. Now using similar arguments as Theorem-1, Section 1.2 we get that $\Tilde{l}_{IMC}$ is rank consistent too. 

\section{Proof of Theorem~2}

We are given that  $\mathbb{E}_{{\Tilde{y}}}[b_i^t-b_{i+1}^t]\geq 0,\;i\in[K-1]$. Let at the $t^{th}$ iteration example $(\x^t,\Tilde{y}^t)$ is being presented to the network. Loss $\Tilde{l}_{CE}$ corresponding to $(\x^t,\Tilde{y}^t)$ is as follows.
\begin{align*}
    &\Tilde{l}_{CE}(g(\x^t),\mathbf{b},\Tilde{y}^t) = \sum_{j=1}^{K}\mathbf{N}^{-1}_{(\Tilde{y}^t,j)}l_{CE}(g(\x^t),\mathbf{b}, j) \\  
    &= - \sum_{j=1}^{K}\mathbf{N}^{-1}_{(\Tilde{y}^t,j)} \sum_{i=1}^{K-1} \left(\log h_i(\x^t)^{z_i^j}+\log(1-h_i(\x^t))^{(1-z_i^j)}\right)
\end{align*}
For every $j=1\ldots K-1$, $z_i^j$ are defined as follows. $z_i^j=1,\;\forall i<j$ and $z_i^j =0,\;\forall i\geq j$.
The update equation using SGD requires to compute the partial derivative of the parameters with respect to the loss function $\Tilde{l}_{CE}$. We see the following.
\begin{align*}
&\frac{\partial \Tilde{l}_{CE}(g(\x^t),\mathbf{b},\Tilde{y}^t)}{\partial b_i} =  -\sum_{j=1}^{K}\mathbf{N}^{-1}_{(\Tilde{y}^t,j)} \Big{[}z_i^j\frac{\partial \log(h_i(\x^t))}{\partial b_i}\\
&\;\;\;+(1-z_i^j)\frac{\partial \log(1-h_i(\x^t))}{\partial b_i}\Big{]}\\
     &= -\sum_{j=1}^{K}\mathbf{N}^{-1}_{(\Tilde{y}^t,j)} \left(\frac{z_i^j}{h_i(\x^t)}-\frac{1-z_i^j}{1-h_i(\x^t)}\right)\frac{\partial h_i(\x^t)}{\partial b_i} \\
     &=-\sum_{j=1}^{K}\mathbf{N}^{-1}_{(\Tilde{y}^t,j)} \Big{(}z_i^j(1-h_i(\x^t))-(1-z_i^j)h_i(\x^t)\Big{)}
\end{align*}
The update equations for thresholds $b_1,\ldots,b_{K-1}$ using SGD are as follows.
Let $\alpha$ be the learning rate. 
\begin{align*}
    b_i^{t+1}&=b_i^t - \alpha \frac{\partial \Tilde{l}_{CE}(g^t(\x^t),\mathbf{b}^t,\Tilde{y}^t)}{\partial b_i}\\
    &=b_i^t + \alpha \sum_{j=1}^{K}\mathbf{N}^{-1}_{(\Tilde{y}^t,j)} \Big{(}z_i^j(1-\sigma(g^t(\x^t)+b_i^t)\\
    &\;\;\;\;\;-(1-z_i^j)\sigma(g^t(\x^t)+b_i^t)\Big{)}
\end{align*}
Using the above equation, we compute the following.
\begin{align*}
&b^{t+1}_i - b^{t+1}_{i+1} = b_i^t - b_{i+1}^t + \alpha \sum_{j=1}^{K}  \mathbf{N}^{-1}_{(\Tilde{y}^t,j)}\Big{(}z_i^j(1-h_i^t(\x^t))\\
&\;\;\;-(1-z_i^j)h_i^t(\x^t)-z_{i+1}^j(1-h_{i+1}^t(\x^t))\\
&\;\;\;+(1-z_{i+1}^j)h_{i+1}^t(\x^t)\Big{)}\\
&= b_i^t - b_{i+1}^t + \alpha \sum_{j=1}^{K}  \mathbf{N}^{-1}_{(\Tilde{y}^t,j)}\Big{[}z_i^j-h_i^t(\x^t)
-z_{i+1}^j\\
&\;\;\;+h_{i+1}^t(\x^t)\Big{]}
\end{align*}
For every $j\in \{1,\ldots,K\}$, there can be three possibilities as follows. (a) $z_i^j=z_{i+1}^j=0$, (b) $z_i^j=z_{i+1}^j=1$ and (c) $z_i^j=1$, $z_{i+1}^j=0$. Thus, we can rewrite $b^{t+1}_i - b^{t+1}_{i+1}$ as follows.

\begin{align*}
&b^{t+1}_i - b^{t+1}_{i+1} = b_i^t - b_{i+1}^t + \alpha \sum_{z_i^j=z_{i+1}^j}  \mathbf{N}^{-1}_{(\Tilde{y}^t,j)}\Big{[}h_{i+1}^t(\x^t)\\
&-h_{i}^t(\x^t)\Big{]}+\alpha \underset{z_i^j=1,z_{i+1}^j=0}{\sum}  \mathbf{N}^{-1}_{(\Tilde{y}^t,j)}\Big{[}1+h_{i+1}^t(\x^t)
-h_{i}^t(\x^t)\Big{]} \\
&= b_i^t - b_{i+1}^t + \alpha \sum_{j=1}^K  \mathbf{N}^{-1}_{(\Tilde{y}^t,j)}\Big{[}h_{i+1}^t(\x^t)
-h_{i}^t(\x^t)\Big{]}\\
&\;\;\;+\alpha   \underset{z_i^j=1,z_{i+1}^j=0}{\sum}  \mathbf{N}^{-1}_{(\Tilde{y}^t,j)}
\end{align*}
Using properties of noise matrix, we know that $\sum_{j=1}^K  \mathbf{N}^{-1}_{(\Tilde{y}^t,j)}=1$. Thus, 
\begin{align*}
b^{t+1}_i - b^{t+1}_{i+1} &= b_i^t - b_{i+1}^t - \alpha \Big{[}h_{i}^t(\x^t)
-h_{i+1}^t(\x^t)\Big{]}\\
&\;\;\;+\alpha   \underset{z_i^j=1,z_{i+1}^j=0}{\sum}  \mathbf{N}^{-1}_{(\Tilde{y}^t,j)}  
\end{align*}
The only possibility for $z_i^j=1,z_{i+1}^j=0$ is $j=i+1$. Thus,
\begin{align*}
 b^{t+1}_i - b^{t+1}_{i+1} &= b_i^t - b_{i+1}^t - \alpha \Big{[}h_{i}^t(\x^t)
-h_{i+1}^t(\x^t)\Big{]}\\
&\;\;\;+\alpha  \mathbf{N}^{-1}_{(\Tilde{y}^t,i+1)}.  
\end{align*}
Since $\mathbf{N}^{-1}_{(\Tilde{y}^t,i+1)}$ updates depend on $\Tilde{y}^t$, we take the expectation on both sides with respect to $\Tilde{y}$, we get the following.
\begin{align}
 \nonumber    \mathbb{E}_{\Tilde{y}}[b^{t+1}_i - b^{t+1}_{i+1}] &\geq  \mathbb{E}_{\Tilde{y}} \Big{[} b_i^t - b_{i+1}^t - \alpha \big{(}h_{i}^t(\x^t)
-h_{i+1}^t(\x^t)\big{)}\Big{]}\\
&\;\; \;\;\;+\alpha \mathbb{E}_{\Tilde{y}}[  \mathbf{N}^{-1}_{(\Tilde{y}^t,i+1)}] \label{eqn:sgd-ce}
    \end{align}
We know that, $h_i^t(\x^t)=\sigma(g^t(\x^t)+b_i^t)$. Also, $b_i^t\geq b_{i+1}^t$. 
Using the Mean-Value Theorem, $\exists \theta \in (b_{i+1}^t,b_i^t)$
such that 
\begin{align*}
\frac{h_i^t(\x^t)-h_{i+1}^t(\x^t)}{b_i^t-b_{i+1}^t}&=\frac{\partial \sigma(g^t(\x^t)+b)}{\partial b}_{\vert \theta}\\ &=\sigma(g^t(\x^t)+\theta)(1-\sigma(g^t(\x^t)+\theta)).
\end{align*}
We know that $0 < \sigma(g^t(\x^t)+\theta)(1-\sigma(g^t(\x^t)+\theta)) \leq 0.25, \; \forall \theta \in \R$. Using this, we get, 
\begin{align*}
    & b_{i}^t-b_{i+1}^t - \alpha(\sigma(g^t(\x^t)+b_i^t)-\sigma(g^t(\x^t)+b_{i+1}^t))\\
    &\;\;= (1-\alpha \frac{\partial \sigma(g^t(\x^t)+b')}{\partial b})(b_{i}^t-b_{i+1}^t)\\
    &\;\;\geq (1-0.25\alpha )(b_{i}^t-b_{i+1}^t)\geq 0
\end{align*}
where the last inequality holds when $\alpha \leq 4$.
Thus for $b_i^t \ge b_{i+1}^t$, we get
\begin{align}
 \label{eqn:MVT}   b_i^t - b_{i+1}^t - \alpha \Big{[}h_{i}(\x^t)
-h_{i+1}(\x^t)\Big{]} \geq 0,\;\forall \alpha \leq 4.
\end{align}
Using eq.(\ref{eqn:sgd-ce}), we know that
\begin{align*}
    \mathbb{E}_{\Tilde{y}}[b^{t+1}_i - b^{t+1}_{i+1}] &\geq  \mathbb{E}_{\Tilde{y}} \Big{[} b_i^t - b_{i+1}^t - \alpha \big{(}h_{i}^t(\x^t)
-h_{i+1}^t(\x^t)\big{)}\Big{]}\\
&\;\; \;\;\;+\alpha \mathbb{E}_{\Tilde{y}}[  \mathbf{N}^{-1}_{(\Tilde{y}^t,i+1)}].
    \end{align*}
    Now, we using the result in eq.(\ref{eqn:MVT}), we get the following.
    \begin{align*}
\mathbb{E}_{\Tilde{y}}[b_i^{t+1}& - b_{i+1}^{t+1}] \geq \alpha  \mathbb{E}_{\Tilde{y}}[\mathbf{N}^{-1}_{(\Tilde{y}^t,i+1)}]\\
 &=\alpha\sum_{k=1}^K P(\Tilde{y}=k) \mathbf{N}^{-1}_{(k,i+1)}\\
  &=\alpha\sum_{k=1}^K \mathbf{N}^{-1}_{(k,i+1)} \sum_{j=1}^K P(y=j)P(\Tilde{y}=k|y=j)  \\
  &=\alpha\sum_{j=1}^K P(y=j)\sum_{k=1}^K  \eta_{(j,k)} \mathbf{N}^{-1}_{(k,i+1)}\\
  &=\alpha\sum_{j=1}^K P(y=j)\mathbb{I}_{\{j=i+1\}}=\alpha P(y=i+1)\geq 0.
\end{align*}
Thus, we have shown that $\mathbb{E}_{\Tilde{y}}[b^{t+1}_i - b^{t+1}_{i+1}] \geq 0$. This completes our proof that SGD gives the optimal solution maintaining rank consistency.

\section{Proof of Theorem~3}
Let at $t^{th}$ iteration example $(\x^t,\Tilde{y}^t)$ is being presented to the network. 
Loss $\Tilde{l}_{IMC}$ corresponding to $(\x^t,\Tilde{y}^t)$ is described as follows.
\begin{align*}
    \Tilde{l}_{IMC}(g(\x^t),\mathbf{b},\Tilde{y}^t) =\sum_{j=1}^K \mathbf{N}^{-1}_{(\Tilde{y}^t,j)} \sum_{i=1}^{K-1} \left[0,1-z_i^j\left(g(\x^t)+b_i\right)\right]_+ 
\end{align*}
Where $z_i^j=1,\;\forall i<j$ and $z_i^j=-1,\;\forall i\geq j$. We first find the sub-gradient of $\Tilde{l}_{IMC}$ w.r.t $b_i$.
\begin{align*}
&\frac{\partial \Tilde{l}_{IMC}(g(\x^t),\mathbf{b},\Tilde{y}^t)}{\partial b_i} =  -\sum_{j=1}^{K}\mathbf{N}^{-1}_{(\Tilde{y}^t,j)} z_i^j \mathbb{I}[z_i^j(g(\x^t)+b_i) < 1] \\
\end{align*}
Hence the SGD based update equation for $b_i$ (with step size $\alpha$) is as follows. 
\begin{align*}
    b_i^{t+1} &= b_i^{t} + \alpha  \sum_{j=1}^{K}\mathbf{N}^{-1}_{(\Tilde{y}^t,j)} z_i^j \mathbb{I}[z_i^j(g^t(\x^t)+b_i^t) < 1] \\
    &=b_i^{t} + \alpha \sum_{j\leq i}\mathbf{N}^{-1}_{(\Tilde{y}^t,j)} z_i^j \mathbb{I}[z_i^j(g^t(\x^t)+b_i^t) < 1]\\
    &\;\;\;\;\;\;+\alpha \sum_{j> i}\mathbf{N}^{-1}_{(\Tilde{y}^t,j)} z_i^j \mathbb{I}[z_i^j(g^t(\x^t)+b_i^t) < 1]\\
    &=b_i^{t} - \alpha \sum_{j\leq i}\mathbf{N}^{-1}_{(\Tilde{y}^t,j)} \mathbb{I}[g^t(\x^t)+b_i^t >-1]\\
    &\;\;\;\;\;\;+\alpha \sum_{j> i}\mathbf{N}^{-1}_{(\Tilde{y}^t,j)}  \mathbb{I}[g^t(\x^t)+b_i^t < 1]
\end{align*}
 Where we used the definition of $z_i^j$. Now, we take the expectation with respect to $\Tilde{y}^t$ on both size, and using the fact that $\mathbb{E}_{\Tilde{y}^t}[\mathbf{N}^{-1}_{(\Tilde{y}^t,j)}]=P(y=j)$, we get the following.
 \begin{align*}
    &\mathbb{E}_{\Tilde{y}^t}[b_i^{t+1} - b_i^{t}]= 
    - \alpha \sum_{j\leq i}\mathbb{E}_{\Tilde{y}^t}[\mathbf{N}^{-1}_{(\Tilde{y}^t,j)}] \mathbb{I}[g^t(\x^t)+b_i^t >-1]\\
    &\;\;\;\;\;\;+\alpha \sum_{j> i}\mathbb{E}_{\Tilde{y}^t}[\mathbf{N}^{-1}_{(\Tilde{y}^t,j)}]  \mathbb{I}[g^t(\x^t)+b_i^t < 1]\\
    &= 
    - \alpha \sum_{j\leq i}P(y=j) \mathbb{I}[g^t(\x^t)+b_i^t >-1]\\
    &\;\;\;\;\;\;+\alpha \sum_{j> i}P(y=j) \mathbb{I}[g^t(\x^t)+b_i^t < 1]\\
    &= 
    - \alpha P(y\leq i) \mathbb{I}[g^t(\x^t)+b_i^t >-1]\\
    &\;\;\;\;\;\;+\alpha P(y>i) \mathbb{I}[g^t(\x^t)+b_i^t < 1]
\end{align*}
Using this, we now compute the following.
\begin{align*}
    & \mathbb{E}_{\Tilde{y}^t}[b_i^{t+1} - b_{i+1}^{t+1} - b_i^{t} + b_{i+1}^{t}] \\
    =&  \mathbb{E}_{\Tilde{y}^t}[b_i^{t+1} - b_i^{t}] - \mathbb{E}_{\Tilde{y}^t}[b_i^{t+1} - b_i^{t}]\\
    =& - \alpha P(y\leq i) \mathbb{I}[g^t(\x^t)+b_i^t >-1]\\
    &\;+\alpha P(y>i) \mathbb{I}[g^t(\x^t)+b_i^t < 1] \\
    &\;+ \alpha P(y\leq i+1) \mathbb{I}[g^t(\x^t)+b_i^t >-1]\\
    &\;-\alpha P(y>i+1) \mathbb{I}[g^t(\x^t)+b_i^t < 1]\\
    =& \alpha [P(y\leq i+1)-P(y\leq i)] \mathbb{I}[g^t(\x^t)+b_i^t >-1]\\
    &\;+\alpha [P(y>i) -P(y>i+1)]\mathbb{I}[g^t(\x^t)+b_i^t < 1]
\end{align*}
But, we know that
\begin{align*}
    P(y>i) -P(y>i+1) \geq 0,\;\forall i \in [K-1]\\
    P(y\leq i+1)-P(y\leq i)\geq 0,\;\forall i \in [K-1]
\end{align*}
and $\mathbb{I}[.]\in \{0,1\}$. Thus, 
\begin{align*}
    &\mathbb{E}_{\Tilde{y}}[(b^{t+1}_{i}-b^{t+1}_{i+1}) - (b^{t}_{i}-b^{t}_{i+1})] \geq 0\\
   \Rightarrow & \mathbb{E}_{\Tilde{y}}[b^{t+1}_{i}-b^{t+1}_{i+1}] \geq \mathbb{E}_{\Tilde{y}}[b^{t}_{i}-b^{t}_{i+1}]=b^{t}_{i}-b^{t}_{i+1}\geq 0
\end{align*} 

This completes the proof. 
\section{Generalisation bounds}

Using unbiased estimator, we have 
\begin{align*}
    \tilde{l}(g(\x), \mathbf{b},y) & = \sum_{j=1}^{K} \mathbf{N}^{-1}_{(y,j)} l(g(\x), \mathbf{b},j) \\
        & = \sum_{j=1}^{K} \mathbf{N}^{-1}_{(y,j)} \sum_{i=1}^{K-1} l^i(g(\x), \mathbf{b},z_i^j) \\
        & = \sum_{i=1}^{K-1} ( \sum_{j=1}^{K} \mathbf{N}^{-1}_{(y,j)}  l^i(g(\x), \mathbf{b},z_i^j) ) \\
        & = \sum_{i=1}^{K-1} \tilde{l}^i(g(\x), \mathbf{b},i)
\end{align*}

For any $i$, if $l^i$  is $L-$Lipschitz, then $\tilde{l}^i$ is $\tilde{L} = (\sum_{j=1}^K |\mathbf{N}^{-1}_{(y,j)}|)L \leq ML$ Lipschitz constant, where $M = \max_y \sum_{j=1}^K |\mathbf{N}^{-1}_{(y,j)}| $.

Using Lipschitz composition property of basic Rademacher generalisation bounds on $i^{th}$ binary classifier, with probability atleast $1-\delta$
\begin{align}\label{gen}
    R_{\tilde{l}^i, \mathit{D}_\rho}(f^i) \leq \hat{R}_{\tilde{l}^i, S}(f^i) + 2ML \mathfrak{R}(\mathcal{F}) + \sqrt{\frac{log(1/\delta)}{2n}}
\end{align}
where $f^i$ is the $i^{th}$ binary classifier. 

Adding the maximal deviations between expected risk and empirical risk for all the $K-1$ classifiers, 
\begin{align*}
    R_{\tilde{l}, \mathit{D}_\rho}(f) \leq \hat{R}_{\tilde{l}, S}(f) + 2ML(K-1) \mathfrak{R}(\mathcal{F}) + \\ (K-1)\sqrt{\frac{log(1/\delta)}{2n}} \numberthis \label{gen1}
\end{align*}
which if true for any $f$. 

Let $$\hat{f} \xleftarrow[]{} \arg \min_{f \in \mathcal{F}} \hat{R}_{\tilde{l},S}(f) $$ and 
$$f^{*} \xleftarrow[]{} \arg \min_{f \in \mathcal{F}} {R}_{l,\mathit{D}}(f)$$
Following Theorem 3 from \cite{Natarajan:2013:LNL:2999611.2999745},
\begin{align*}
   & R_{l,\mathit{D}}(\hat{f}) - R_{l,\mathit{D}}(f^{*}) = R_{\tilde{l},\mathit{D}_\rho}(\hat{f}) - R_{\tilde{l},\mathit{D}_\rho}(f^{*}) \\
   & = \hat{R}_{\tilde{l},S}(\hat{f}) - \hat{R}_{\tilde{l},S}(f^*) +  (R_{\tilde{l},\mathit{D}_\rho}(\hat{f}) - \hat{R}_{\tilde{l},S}(\hat{f})) \\ & \quad  + (\hat{R}_{\tilde{l},S}(f^*) - R_{\tilde{l},\mathit{D}_\rho}(f^{*})  )\\
    & \leq 2 \max_{f\in \mathcal{F}} |R_{\Tilde{l},\mathit{D}_\rho}(f)-\hat{R}_{\Tilde{l},S}(f)|  \numberthis \label{gen2}
\end{align*}

From \ref{gen1} and \ref{gen2}, we get, 
\begin{align*}
R_{l,\mathit{D}}(\hat{f})  \leq R_{l,\mathit{D}}(f^{*}) + 4ML(K-1) \mathfrak{R}(\mathcal{F}) + \\ 2(K-1)\sqrt{\frac{log(1/\delta)}{2n}}
\end{align*}

\section{Noise Matrix}
We give an sample noise matrix here for california housing dataset. 

Actual noise matrix = $\begin{bmatrix}
        0.725 & 0.15 & 0.075 & 0.05 \\
       0.15 & 0.625 & 0.15 & 0.075 \\
       0.075 & 0.15 & 0.625 & 0.15 \\
       0.05 & 0.075 & 0.15 & 0.725
\end{bmatrix}$

Estimated Matrix = $\begin{bmatrix}
        0.80&0.14&0.03&0.02 \\ 
        0.1&0.64&0.15&0.06 \\ 
        0.03&0.11&0.63&0.23 \\ 
        0.02&0.11&0.18&0.69
\end{bmatrix}$

\bibliographystyle{plain} \bibliography{v_shape}

\begin{thebibliography}{10}

\bibitem{Antoniuk2016}
Kostiantyn Antoniuk, Vojt{\v{e}}ch Franc, and V{\'a}clav Hlav{\'a}{\v{c}}.
\newblock V-shaped interval insensitive loss for ordinal classification.
\newblock {\em Machine Learning}, 103(2):261--283, May 2016.

\bibitem{DBLP:journals/corr/abs-1901-07884}
Wenzhi Cao, Vahid Mirjalili, and Sebastian Raschka.
\newblock Consistent rank logits for ordinal regression with convolutional
  neural networks.
\newblock {\em CoRR}, abs/1901.07884, 2019.

\bibitem{DBLP:nn_based_approach}
Jianlin Cheng.
\newblock A neural network approach to ordinal regression.
\newblock {\em CoRR}, abs/0704.1028, 2007.

\bibitem{Chu2005NewAT}
Wei Chu and S.~Sathiya Keerthi.
\newblock New approaches to support vector ordinal regression.
\newblock In {\em ICML}, 2005.

\bibitem{Crammer:2001:PR:2980539.2980623}
Koby Crammer and Yoram Singer.
\newblock Pranking with ranking.
\newblock In {\em NIPS}, pages 641--647, 2001.

\bibitem{alzeimhers}
Orla~M. Doyle, Eric Westman, Andre~F. Marquand, Patrizia Mecocci, Bruno Vellas,
  Magda Tsolaki, Iwona Kłoszewska, Hilkka Soininen, Simon Lovestone, Steve
  C.~R. Williams, and Andrew Simmons.
\newblock Predicting progression of alzheimer’s disease using ordinal
  regression.
\newblock {\em PLOS ONE}, 9(8):1--10, 08 2014.

\bibitem{noise_survey}
B.~{Frenay} and M.~{Verleysen}.
\newblock Classification in the presence of label noise: A survey.
\newblock {\em IEEE Transactions on Neural Networks and Learning Systems},
  25(5):845--869, 2014.

\bibitem{ghosh2017robust}
Aritra Ghosh, Himanshu Kumar, and PS~Sastry.
\newblock Robust loss functions under label noise for deep neural networks.
\newblock In {\em Thirty-First AAAI Conference on Artificial Intelligence},
  2017.

\bibitem{credit_ratings}
Rainer Hirk, Kurt Hornik, and Laura Vana.
\newblock Multivariate ordinal regression models: an analysis of corporate
  credit ratings.
\newblock {\em Statistical Methods {\&} Applications}, Aug 2018.

\bibitem{Horn:2012:MA:2422911}
Roger~A. Horn and Charles~R. Johnson.
\newblock {\em Matrix Analysis}.
\newblock Cambridge University Press, New York, NY, USA, 2nd edition, 2012.

\bibitem{diseases}
Shivalingappa. Javali and Parameshwar. Pandit.
\newblock {A comparison of ordinal regression models in an analysis of factors
  associated with periodontal disease}.
\newblock {\em Journal of Indian Society of Periodontology}, 14(3):155--159,
  2010.

\bibitem{article_Adamw}
Diederik Kingma and Jimmy Ba.
\newblock Adam: A method for stochastic optimization.
\newblock {\em International Conference on Learning Representations}, 12 2014.

\bibitem{Li:2006:ORE:2976456.2976565}
Ling Li and Hsuan-Tien Lin.
\newblock Ordinal regression by extended binary classification.
\newblock In {\em Proceedings of the 19th International Conference on Neural
  Information Processing Systems}, NIPS'06, pages 865--872, Cambridge, MA, USA,
  2006. MIT Press.

\bibitem{7159100}
T.~{Liu} and D.~{Tao}.
\newblock Classification with noisy labels by importance reweighting.
\newblock {\em IEEE Transactions on Pattern Analysis and Machine Intelligence},
  38(3):447--461, March 2016.

\bibitem{Liu2018ACD}
Yanzhu Liu, Adams Wai-Kin Kong, and Chi~Keong Goh.
\newblock A constrained deep neural network for ordinal regression.
\newblock {\em CVPR}, pages 831--839, 2018.

\bibitem{DBLP:pril/corr/abs-1802-03873}
Naresh Manwani.
\newblock {PRIL: Perceptron Ranking Using Interval Labeled Data}.
\newblock In {\em {CoDS-COMAD}}, pages 78--85, Kolkata, India, 2019.

\bibitem{DBLP:noise_tolerance_nar}
Naresh Manwani and P.~S. Sastry.
\newblock Noise tolerance under risk minimization.
\newblock {\em {IEEE} Trans. Cybernetics}, 43(3):1146--1151, 2013.

\bibitem{Natarajan:2013:LNL:2999611.2999745}
Nagarajan Natarajan, Inderjit~S. Dhillon, Pradeep Ravikumar, and Ambuj Tewari.
\newblock Learning with noisy labels.
\newblock In {\em NIPS}, pages 1196--1204, 2013.

\bibitem{Patrini_2017_CVPR}
Giorgio Patrini, Alessandro Rozza, Aditya Krishna~Menon, Richard Nock, and
  Lizhen Qu.
\newblock Making deep neural networks robust to label noise: A loss correction
  approach.
\newblock In {\em {CVPR}}, July 2017.

\bibitem{decoding_brain}
Emi Satake, Kei Majima, Shuntaro~C. Aoki, and Yukiyasu Kamitani.
\newblock Sparse ordinal logistic regression and its application to brain
  decoding.
\newblock {\em Frontiers in Neuroinformatics}, 12:51, 2018.

\end{thebibliography}

\end{document}